\documentclass[runningheads]{llncs}

\usepackage[mobile]{eccv}

\usepackage{eccvabbrv}
\usepackage{microtype}
\usepackage{graphicx}
\usepackage{booktabs}
\usepackage{caption}
\usepackage[accsupp]{axessibility}  % Improves PDF readability for those with disabilities.

\usepackage{hyperref}

\usepackage{orcidlink}
\usepackage{comment}
\begin{document}

\title{SparseCraft: Few-Shot Neural Reconstruction through Stereopsis Guided Geometric Linearization} 

\titlerunning{SparseCraft}

%\author{Mae Younes\inst{1}$^*$\orcidlink{0000-0002-4831-3343} \and
%Amine Ouasfi\inst{1,2}$^*$ \and Adnane Boukhayma\inst{2} 
%}

\author{Mae Younes$^{\dagger *}$\orcidlink{0000-0002-4831-3343} \and
Amine Ouasfi$^{*}$ \and Adnane Boukhayma}

\authorrunning{M.~Younes et al.}

\institute{Inria, $^{\dagger}$Univ. Rennes, CNRS, IRISA, M2S, France}

\maketitle
\def\thefootnote{*}\footnotetext{These authors contributed equally to this work}\def\thefootnote{\arabic{footnote}}

\begin{abstract}
  We present a novel approach for recovering 3D shape and view dependent appearance from a few colored images, enabling efficient 3D reconstruction and novel view synthesis. Our method learns an implicit neural representation in the form of a Signed Distance Function (SDF) and a radiance field. The model is trained progressively through ray marching enabled volumetric rendering, and regularized with learning-free multi-view stereo (MVS) cues. Key to our contribution is a novel implicit neural shape function learning strategy that encourages our SDF field to be as linear as possible near the level-set, hence robustifying the training against noise emanating from the supervision and regularization signals. Without using any pretrained priors, our method, called SparseCraft, achieves state-of-the-art performances both in novel-view synthesis and reconstruction from sparse views in standard benchmarks, while requiring less than 10 minutes for training. Project page: \href{https://sparsecraft.github.io}{sparsecraft.github.io}

\end{abstract}
\section{Introduction}
\label{sec:intro}
Replicating the 3D world around us digitally in a faithful manner is a long-standing problem that has prompted substantive research in computer vision and graphics alike, with countless downstream applications. While current solutions can provide impressive results, many of them still rely on abundantly informative input, be it in quality (\eg high resolution imagery, depth sensors) or quantity (\eg dense arrays of views). However, due to many constrained scenarios (\eg out-of-the-studio, low budget, \etc) and in the interest of wider applicability, the community (\eg \cite{ren2022volrecon,long2022sparseneus,johari2022geonerf,yang2023freenerf,niemeyer2022regnerf,NeuralTPS,ouasfi2024unsupervised,ouasfi2024fewshot}) is actively seeking solutions that can deliver under minimal input. 

Given a few colored images, we aim to capture both the shape and appearance of the observed object or scene. In practice, we seek metrically accurate 3D reconstruction, and photo-realistic novel view synthesis. In this regard, traditional computational photogrammetry combines structure from motion (SfM) and multi-view stereo (MVS)\cite{schonberger2016structure,schoenberger2016mvs} to  provide calibration and triangulate an explicit geometry based on matching. However, it can lead to noisy and incomplete meshes in challenging and non Lambertian scenarios. On the other hand, deep learning based implicit neural representations (INR) have emerged as a powerful tool for 3D modelling~\cite{mildenhall2020nerf,wang2021neus}. They have shown ability to learn both detailed shape and radiance though image supervised differentiable volumetric rendering from dense image arrays~\cite{wang2021neus,yariv2021volume}. Learning such accurate implicit shape representations remains challenging when only a few images are available. Current methods in the literature~\cite{ren2022volrecon,long2022sparseneus,liang2023retr} rely on learned data priors across many training scenes, by conditioning the implicit representation on spatially local features obtained from the sparse input images through generalizable encoders. 
These can suffer nonetheless from out of distribution generalization issues, and typically require substantial and expensive calibrated multi-view data for training.

Differently, we advocate to fit a neural signed distance (SDF) and radiance functions self-supervisedly to the images. Using a progressively learned hash encoding~\cite{muller2022instant} provides regularization and a more stable and efficient training. We use MVS geometry and color cues to further regularize this challenging learning task.  Notice that these cues are readily available, as photogrammetry is typically required to obtain the calibration needed for learning INRs. 

Unfortunately, our training is facing noisy labels or phenomena that can be interpreted as such: \eg The MVS geometry can be noisy, and the volumetric rendering supervision can be imprecise due to imperfect calibration, and the inherent bias of geometry based volumetric rendering~\cite{fu2022geoneus}. To alleviate these challenges, we propose a novel loss rooted in Occam’s razor principle. We focus on the surface \ie near the MVS samples, as it is the most critical region in our learning.  We hypothesize that excess non-linearity~\cite{srinivas2022efficient} there can lead to overfitting on the noise. Hence, we encourage our SDF to be as linear as possible near the MVS samples, by making the function approximate its first order Taylor expansion, and we integrate the MVS point and normal supervision in this linearization (Section~\ref{sec:reg}). We show empirically that this loss leads to considerable improvement in our method, as compared to the previous methods, and also a directly MVS supervised baseline.

We obtain state-of-the-art performances in both novel view synthesis and reconstruction using standard metrics, as-well-as superior qualitative results to previous methods, without using any pre-learned priors, and within shorter training times. In summary, our main contributions are:\\
\noindent$\bullet$ Few-shot 3D reconstruction and novel view synthesis without any pre-learned data priors.\\
\noindent$\bullet$ Leveraging a progressive multi-resolution hash learning strategy in this context.\\
\noindent$\bullet$ A framework, that we call SparseCraft, for harnessing all MVS data: points, normals to regularize the SDF, and color to regularize the diffuse.\\
\noindent$\bullet$ Our novel Taylor expansion inspired losses to regularize SDF learning from sparse multi-view imagery.

\section{Related Work}
\noindent\textbf{Multi-View Stereo}
Conventional MVS can be classified based on scene representation: volumetric~\cite{kutulakos2000theory, seitz1999photorealistic, kostrikov2014probabilistic}, point cloud ~\cite{lhuillier2005quasi, furukawa2010}, and depth map based~\cite{galliani_2015_gipuma, schoenberger2016mvs, xu_2019_acmm}. Recently, there has been a preference for depth map based methods due to their versatility, dividing the problem into depth map estimation and fusion stages~\cite{galliani_2015_gipuma, schoenberger2016mvs}. Subsequently, Poisson reconstruction~\cite{kazhdan2013screened} is applied to the fused point cloud to produce a watertight mesh. 
Despite significant advancements, extreme scenarios such as minimal input can still prove challenging for such methods, often resulting in incomplete and inaccurate reconstructions. Nevertheless, we demonstrate that by judiciously leveraging the incomplete fused point cloud generated from one such method (COLMAP~\cite{schonberger2016structure}), our approach surpasses the state-of-the-art in multi-view few-shot reconstruction.

\noindent\textbf{Implicit Neural Representations}
Implicit Neural fields employ deep neural networks to model 2D or 3D data as continuous functions, overcoming many of the limitations of explicit ones
(\eg meshes \cite{wang2018pixel2mesh,kato2018neural,jena2022neural} and point clouds \cite{fan2017point,aliev2020neural,kerbl20233d})
in modelling shape, radiance and light fields (\eg \cite{mildenhall2020nerf,yariv2021volume,wang2021neus,jain2021dreamfields,chan2022efficient,li2023learning,li2023regularizing}).
The seminal work (NeRF)~\cite{mildenhall2020nerf}, which combines volume rendering and implicit representations, has paved the way for learning  diverse tasks, including novel view synthesis~\cite{mildenhall2020nerf, martin2021nerf}, 3D generation~\cite{poole2022dreamfusion, jain2022zero}, deformation~\cite{park2021nerfies, pumarola2021d, Rebain20arxiv_derf}, and video rendering~\cite{Li20arxiv_nsff, Xian20arxiv_stnif, Du20arxiv_nerflow, li2022neural}. More recently, attention was shed on implicit surface reconstruction, with a focus on single-stage optimization and robust representation potential~\cite{idr_yariv2020multiview, yang2022neumesh}. This was improved through novel weight functions involving SDFs for color accumulation during volumetric rendering~\cite{wang2021neus, yariv2021volume}. However, persistent challenges such as geometric bias arising from discrete sampling and other factors~\cite{fu2022geoneus, zhang2023towards} still remain.

Efforts have been directed towards addressing the time-intensive training associated with these methods. \ie, NeRF-based work~\cite{yu_and_fridovichkeil2021plenoxels, SunSC22, muller2022instant} introduced voxel-grid features. Subsequently, other literature~\cite{Voxurf, Wang_2023_ICCV} extended them to surfaces. Lately, Neuralangelo~\cite{li2023neuralangelo} proposed leveraging multi-resolution hash grids with numerical gradient computation and a topology warm-up strategy for neural surface reconstruction. While achieving high-fidelity geometry from dense images, it comes with a considerable training time cost. Inspired by the latter, our work leverages numerical gradients and hash encoding, and additionally employs an occupancy grid for sampling, to strike a balance between reconstruction quality and training speed. 

\noindent\textbf{Novel-View from Sparse Input}
Existing work has tackled this task by incorporating additional information, such as normalization-flow~\cite{niemeyer2022regnerf}, perceptual~\cite{zhang2021ners} and diffusion-based~\cite{wynn2023diffusionerf} regularization, depth supervision~\cite{roessle2022dense, deng2022depth, wei2021nerfingmvs}, and enforcing cross-view semantic consistency~\cite{jain2021putting}. Conversely, another line of work~\cite{chibane2021stereo, yu2021pixelnerf, chen2021mvsnerf} strives to develop transferable models by training on a large, curated dataset, eschewing the use of external models. Recent investigations posit that geometry is a pivotal factor in few-shot neural rendering, advocating for geometry regularization~\cite{niemeyer2022regnerf} to enhance performance. However, these methods require resource-intensive pre-training on tailored multi-view datasets~\cite{chibane2021stereo, yu2021pixelnerf, chen2021mvsnerf} or employing costly training-time patch rendering~\cite{jain2021putting, niemeyer2022regnerf}, thus introducing significant overhead.

Conversely, other approaches propose regularization strategies during single scene fitting. These include frequency encoding regularization~\cite{yang2023freenerf}, entropy constraints on density~\cite{Kim_2022_CVPR}, utilizing a mixture density model~\cite{Seo_2023_CVPR} or exploiting flipped reflection rays as augmentation~\cite{Seo_2023_ICCV}. In this work, we demonstrate that explicit regularization, aimed at linearizing the signed distance function in proximity to the surface with guidance from an incomplete point cloud derived from a classical MVS method, along with the incorporation of a progressive hash encoding, not only enhances the surface reconstruction of our SDF based method, but also enables it to surpass state-of-the-art NeRF-based approaches in rendering quality in the few-shot object-centric setting.

\noindent\textbf{Reconstruction from Sparse Input}
For this task, geometric priors~\cite{fu2022geoneus, zhang2022critical, wang2022neuris, yu2022monosdf, ouasfi2024robustifying} have been proposed to enhance reconstruction in the single scene fitting setting. However, these methods are slow to train and still display artifacts and failures. Generalizable novel view synthesis models ~\cite{yu2021pixelnerf, chen2021mvsnerf, wang2021ibrnet, liu2022neuray, trevithick2021grf, sitzmann2019scene, peng2020convolutional} can be repurposed for reconstruction by carefully adjusting the density threshold for extraction. However, their reconstructions tend to be noisy and not as robust as reconstruction methods. Generalizable surface reconstruction methods (from images ~\cite{long2022sparseneus, ren2022volrecon} as well as point clouds ~\cite{ouasfi2024Mixing,ouasfi2022few,boulch2022poco,peng2020convolutional}) are still prone to failure for out-of-distribution scenes/views. Another noteworthy work~\cite{wu2023s} employs the neural rendering of an implicit reconstruction method to improve the MVS performance of deep MVS models in the few-shot setting. In contrast to all the aforementioned work, our rapidly trained method achieves state-of-the-art results for the few-shot reconstruction task without relying on pre-learned priors.

\section{Method}
\label{sec:method}
\begin{figure}[t!]
\centering
\includegraphics[width=0.8\linewidth]{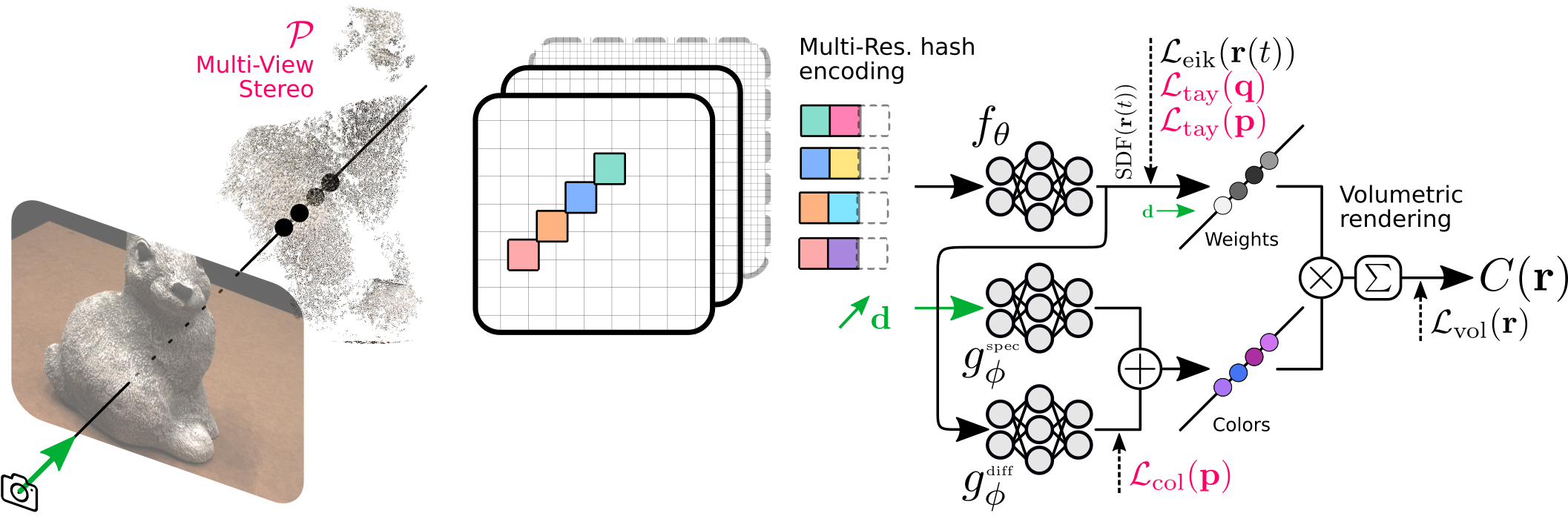}
\caption{\small Overview: In this toy example, we illustrate inference given 4 samples $\{\mathbf{r}(t)\}$ on a ray $\mathbf{r}$ (where the last hash resolution is not active yet). Dashed arrows symbolize losses operating mid-training. SparseCraft leverages differentiable volumetric rendering to learn a SDF based implicit representation given a few images, using MVS cues as regularization ( losses in Red). }
\label{fig:pipe}
\end{figure}

Given a few input colored images $\{I_i\}_{i=1}^{N}$, our goal is to recover the shape and appearance of the observation. We achieve this by learning implicit shape and radiance functions simultaneously. 
We model the shape $\mathcal{S}$ with a SDF $f$ parameterized with a neural network $f_\theta$. We model the radiance as a view direction $\mathbf{d}$ and location $\mathbf{x}$ dependent 3-channel color function $g$ parameterized through a neural network $g_\phi$, \ie $g(\mathbf{x},\mathbf{d})=\mathbf{c}$ where $\mathbf{c}\in[0,1]^{3}$. The inferred shape $\hat{\mathcal{S}}$ can be obtained at test time as the zero level set of the learned SDF $f_\theta$ at convergence: 
$\hat{\mathcal{S}} = \{\mathbf{x}\in\mathbb{R}^3 \mid f_\theta(\mathbf{x}) = 0\}.$ 
Concurrently, given a target new view point, a novel image $\hat{I}$ can be generated through ray-wise volumetric rendering~\cite{kajiya1984ray} per pixel, using the converged neural SDF and radiance fields $f_\theta$ and $g_\phi$ respectively. 

\subsection{Learning Implicit Neural Shape and Radiance}
The NeRF~\cite{mildenhall2020nerf} framework enables learning a volumetric scene representation through a synthesis and compare procedure between generated and ground-truth pixel values. Let us assume a ray $\mathbf{r}(t) = \mathbf{o} + t\mathbf{d}$, where $\mathbf{o}$ is the camera origin and $\mathbf{d}$ the ray direction. The color $C$ of the pixel corresponding to a ray $\mathbf{r}$ can be generated through integration along the ray:   
\begin{align} 
C(\mathbf{r}) &= \int T(t)\sigma(\mathbf{r}(t))\mathbf{c}(\mathbf{r}(t),\mathbf{d})dt.
\label{equ:volrend}
\end{align}
where $\sigma(\mathbf{r}(t))$ denotes volume density, which represents a differential opacity signaling the amount of radiance  accumulated by a ray passing through the point $\mathbf{r}(t)$. $T(t)$ denotes transparency, \ie the accumulated transmittance along the ray until $t$, which can be derived from density accordingly:
\begin{equation}
T(t) = \exp\left(-\int^{t}\sigma(\mathbf{r}(s)) ds\right).
\label{equ:trans}
\end{equation}
As extracting geometry by thresholding density $\sigma$ yields suboptimal and noisy results, recent literature proposed to involve a SDF $f$ in the volumetric rendering equation, by defining a function $\Psi$ that transforms the SDF into density $\sigma(\mathbf{r}(t))$ for Yariv \etal~\cite{yariv2021volume}, the weighting function $T(t)\sigma(\mathbf{r}(t))$ in the case of Wang \etal~\cite{wang2021neus}, and most recently the transparency $T(t)$ in the work by Wang \etal~\cite{wang2022hf}. We follow here the latter representation, where the transformation $\Psi$ is chosen to satisfy monotony and boundary conditions fit for $T(t)$:
\begin{align}
T(t)&=\Psi_s(f(\mathbf{r}(t))) = \frac{1}{1+\exp({-sf(\mathbf{r}(t))})},
\end{align}
where $s$ controls the slope of the transformation. 
In practice, the integral in Equation~\ref{equ:volrend} is approximated using discrete samples $\{t_i\}$ with the quadrature rule~\cite{max1995optical}. Using our SDF and radiance neural networks $f_{\theta}$ and $g_{\phi}$, the inferred color of a ray then writes:
\begin{equation}
C(\mathbf{r}) = \sum T_i \left(1-e^{-\sigma_i(t_{i+1}-t_i)}\right)g_{\phi}(\mathbf{r}(t_i),\mathbf{d}),
\label{equ:disc}
\end{equation}
where the transparency and density are obtained from the SDF network~\cite{wang2022hf}. 
Model parameters $\theta$ and  $\phi$ can be optimized at this stage using the following empirical risk minimization:
\begin{equation}
\min_{\theta,\phi}\underset{\begin{subarray}{c}
\mathbf{r}\sim \mathcal{R}\\
t\sim \mathcal{T}_{\mathbf{r}}
\end{subarray}}{\mathbb{E}\ } \mathcal{L}_{\text{vol}}(\mathbf{r}) +  \mathcal{L}_{\text{eik}}(\mathbf{r}(t)),
\end{equation}
where $\mathcal{R}$ symbolizes a distribution over training rays among all training images, and $\mathcal{T}_{\mathbf{r}}$ is the sample distribution along a ray $\mathbf{r}$. $\mathcal{L}_{\text{vol}}$ is the photometric reconstruction loss based on discretized volumetric rendering (Equation~\ref{equ:disc}), while $\mathcal{L}_{\text{eik}}$ is the Eikonal regularization~\cite{gropp2020implicit} that helps avoid the all zero SDF degenerate solution: 
\begin{gather}
\mathcal{L}_{\text{vol}}(\mathbf{r}) = ||C(\mathbf{r})-C_{gt}(\mathbf{r})||_1.\\
\mathcal{L}_{\text{eik}}(\mathbf{r}(t)) = \left(||\nabla f(\mathbf{r}(t))||_2 - 1 \right)^2.
\end{gather}
\subsection{Regularization with Stereopsis Cues}
\label{sec:reg}

Learning a SDF and radiance field conjointly from few images can be an under-constrained problem, which underpins the need for additional regularization. Fitting 3D INRs to images entails typically an automatic calibration preprocess. The latter estimates camera intrinsic and extrinsic parameters, which are key to performing 3D consistent volumetric rendering and/or ray marching. The main method of choice in this context remains COLMAP\cite{schonberger2016structure,schoenberger2016mvs} (SfM + MVS). Hence, without additional overhead, we can acquire the dense fused MVS point cloud, with its point-wise normal estimations and color labels using the sparse input images. We propose subsequently to regularize our training with these additional cues. We note that while MVS points have been exploited before in learning NeRFs from sparse~\cite{deng2022depth,roessle2022dense,wei2021nerfingmvs} and dense images~\cite{xu2022point}, and SDF based radiance from dense images~\cite{johari2022geonerf}, we propose differently here to use these cues in learning SDF based radiance in the sparse setting. Additionally, and to the best of our knowledge, our work is also the first to suggest leveraging the color and normal MVS labels, and not only the point spatial locations. 

However, the MVS surface samples come with a considerable deal of noise, while also being incomplete, due to inaccuracies in the matching and triangulation process that further intensify in our sparse input setting, along with MVS related limitations when dealing with challenging surfaces (\eg textureless and reflective surfaces).  Furthermore, we can argue that even our volumetric rendering based supervision is prone to noise. As a matter of fact, this noise can be manifested in \eg 3D inconsistent supervision emanating from imprecision in the calibration, and also in the inherent bias~\cite{xu2022point} arising from volumetric integration based geometry modeling, as opposed to \eg  a pinpoint root rasterization based geometry modelling. We propose a novel strategy to remedy these challenges in the following.

\noindent\textbf{Taylor Expansion Based Geometric Regularization}
We focus on the level set of our SDF, where the most crucial knowledge for rendering concentrates. We hypothesize that encouraging our SDF to be as linear as possible there can robustify it against the noise introduced above as intuitively, overly complex models are more likely to overfit on noisy samples~\cite{qin2019adversarial}.
We derive a loss that can achieve this linearization efficiently, while integrating MVS point and normal label supervision seamlessly. 

We denote by $\mathcal{P} \subset \mathbb{R}^{3\times M}$ the MVS  point cloud obtained from input images $\{I_i\}$. We note that each sample $\mathbf{p}\in \mathcal{P}$ comes with a normal $\vec{\mathbf{n}}_{\text{\tiny MVS}}(\mathbf{p})$ and color $\mathbf{c}_{\text{\tiny MVS}}(\mathbf{p})$ estimation. We generate a pool of query points near the surface by sampling around the MVS points following a normal distribution, \ie $\{\mathbf{q} \sim \mathcal{N}(\mathbf{p},\sigma_{\epsilon} \mathbf{I}_3)\}$ where the standard deviation $\sigma_{\epsilon}$ decreases proportionately with the progressive learning step $\epsilon$ during training. We recompute subsequently the nearest point $\mathbf{p}$ in $P$ for each sample $\mathbf{q}$, thus forming the following set of training pairs:
\begin{equation}
\mathcal{Q} := \{(\mathbf{q},\mathbf{p}), \mathbf{p}=\min_{\mathbf{v}\in\mathcal{P}}||\mathbf{v}-\mathbf{q}||_2\}.
\end{equation}

Given a pair $(\mathbf{q},\mathbf{p})$ in $\mathcal{Q}$, let us consider the first order Taylor polynomial approximation of our SDF $f_{\theta}$ around $\mathbf{q}$, and evaluate it at $\mathbf{p}$, secure in the knowledge that it is in the direct vicinity of $\mathbf{q}$: 
\begin{equation}
f_{\theta}(\mathbf{p})\approx f_{\theta}(\mathbf{q}) + \nabla f_{\theta}(\mathbf{q})^\top(\mathbf{p}-\mathbf{q}).
\label{equ:tay1}
\end{equation}
Leveraging the constraint that points $\mathbf{p}$ need to belong to the zero level set, we can derive the following loss:
\begin{equation}
\mathcal{L}_{\text{tay}}(\mathbf{q}) = ||f_{\theta}(\mathbf{q})+\nabla f_{\theta}(\mathbf{q})^\top(\mathbf{p}-\mathbf{q})||_2.
\end{equation}

Note that this loss encourages both our function to have minimal curvature near the surface, and MVS points to coincide with the level set. 

Multiplying by gradient $\nabla f_{\theta}$ and rearranging Equation~\ref{equ:tay1} leads to the following approximation:
\begin{equation}
\mathbf{p} \approx \mathbf{q} - f_{\theta}(\mathbf{q})\cdot 
\frac{\nabla f_{\theta}(\mathbf{q})}{||\nabla f_{\theta}(\mathbf{q})||_2^2}.
\end{equation}
Hence, our loss $\mathcal{L}_{\text{tay}}(\mathbf{q})$ can also be interpreted as supervising a single step of Newton root finding  on the SDF $f_{\theta}$, initialized at $\mathbf{q}$, with its nearest neighbor in the MVS point cloud $\mathbf{p}$.

Let us consider now the Taylor approximation of our SDF around $\mathbf{p}$ conversely, as evaluated at query $\mathbf{q}$:
\begin{equation}
f_{\theta}(\mathbf{q})\approx f_{\theta}(\mathbf{p}) + \nabla f_{\theta}(\mathbf{p})^\top(\mathbf{q}-\mathbf{p}).
\end{equation}
We can leverage here the additional constraint that the normalized gradient of the SDF needs to approximate the surface normal, \ie $\vec{\mathbf{n}}_{\text{\tiny MVS}}(\mathbf{p}) \approx \nabla f_\theta(\mathbf{p})/||\nabla f_\theta(\mathbf{p})||_2$ . Thus, we can derive the loss:
\begin{equation}
\mathcal{L}_{\text{tay}}(\mathbf{p}) = ||f_{\theta}(\mathbf{q})-||\nabla f_\theta(\mathbf{p})||_2\cdot \vec{\mathbf{n}}_{\text{\tiny MVS}}(\mathbf{p})^\top(\mathbf{q}-\mathbf{p})||_2.
\label{equ:tay2}
\end{equation}
Table~\ref{tab:taylor_chamfer} and Figure~\ref{fig:taylor_ablation} show the benefit of using these Taylor losses as opposed to standard direct supervision.   

\noindent\textbf{Color Regularization}
The color labels provided by MVS are averaged from the input images, so we propose to use them as a supervision to the diffuse component of our radiance. Hence, as illustrated in Figure~\ref{fig:pipe}, and following~\cite{li2023neuralangelo}, our color network consists of two small MLPs $g_{\phi}^{\text{\tiny diff}}$ and $g_{\phi}^{\text{\tiny spec}}$ modelling view independent and view dependent radiance respectively:
\begin{align}
g_{\phi}(\mathbf{x},\mathbf{d}) &:= g_{\phi}^{\text{\tiny spec}}(\mathbf{x},\mathbf{F}_{\theta}(\mathbf{x}),\nabla f_\theta(\mathbf{x})/||\nabla f_\theta(\mathbf{x})||_2,\mathbf{d})
+ g_{\phi}^{\text{\tiny diff}}(\mathbf{x},\mathbf{F}_{\theta}(\mathbf{x})),
\end{align}
where $\mathbf{F}_{\theta}$ is a feature extracted from the geometry network $f_\theta$. Our color regularization applied at MVS points then writes: 
\begin{equation}
\mathcal{L}_{\text{col}}(\mathbf{p}) = ||g_{\phi}^{\text{\tiny diff}}(\mathbf{p})-\mathbf{c}_{\text{\tiny MVS}}(\mathbf{p})||_1.
\end{equation}

Finally, we can learn our implicit neural representation through the following combined optimization:
\begin{equation}
\min_{\theta,\phi}\underset{\begin{subarray}{c}
\mathbf{r}\sim \mathcal{R}\\
t\sim \mathcal{T}_{\mathbf{r}}\\
(\mathbf{q},\mathbf{p}) \sim \mathcal{Q}
\end{subarray}}{\mathbb{E}\ } \mathcal{L}_{\text{vol}}(\mathbf{r}) +  \mathcal{L}_{\text{eik}}(\mathbf{r}(t)) + \mathcal{L}_{\text{tay}}(\mathbf{q}) + \mathcal{L}_{\text{tay}}(\mathbf{p}) + \mathcal{L}_{\text{col}}(\mathbf{p}).
\end{equation}

Figure~\ref{fig:pipe} provides a visual summary of our method.
\subsection{Fast Progressive Learning}
Our implementation of the INR network builds on the seminal work in~\cite{muller2022instant}. Our SDF $f_{\theta}$ consists of an efficiently CUDA implemented multi-resolution hash encoding followed by a small MLP, and the radiance network $g_{\phi}$ consists of two small MLPs. We use an explicit occupancy grid that guides the sampling along rays (\ie $t\sim \mathcal{T}_{\mathbf{r}}$) for inference. This combination allows for fast training.

While progressive learning through positional encoding or learnable features was introduced previously for learning NeRFs from dense (\eg~\cite{lin2021barf}) and sparse (\eg~\cite{yang2023freenerf}) images, and SDF based radiance from dense images (\eg~\cite{li2023neuralangelo}), we propose here to explore this strategy for SDF based radiance learning in the sparse setting for the first time to the best of our knowledge. Differently from Neuralangelo~\cite{li2023neuralangelo}, we use the progressive hash encoding to regularize the training in the few shot setting. Hence, it is applied throughout the whole training, rather than its use as a warm-up strategy in~\cite{li2023neuralangelo}. We note that progressively activating hash resolutions during training reduces overfitting and improves the stability of the training in our experiments, and also improves the rendering quality. We also use numerical gradient to approximate derivatives, which allows to back-propagate gradients to more hash cells in the training. 
The step size of the derivative computation $\epsilon$ is scheduled progressively in concordance with the hash dimensions, as recommended in~\cite{li2023neuralangelo}.
Details of the scheduling of our progressive learning are reported in the supplementary material. 

\section{Implementation Details}
\label{sec:imp}
We build upon the instant-nsr-pl~\cite{instant-nsr-pl} implementation of Neuralangelo and utilize Nerfacc's~\cite{li2023nerfacc} accelerated sampling with occupancy grid. Our hash resolution spans from $2^2$ to $2^{11}$ with 32 levels, and we employ a multi-level optimization strategy. We use AdamW optimizer with a learning rate schedule and a combination of losses with varying weights. For more details on the implementation, including the architecture of our MLPs, training protocol, and cues sampling, please refer to the supplementary material.

\section{Experiments}
\label{sec:experiments}

\begin{figure}[t!]
    \centering
    \includegraphics[width=0.7\linewidth]{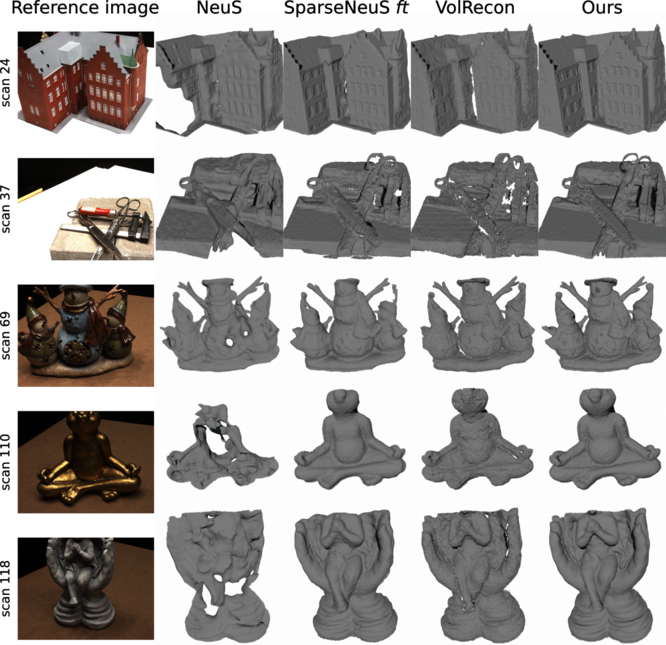}
    \caption{Qualitative comparison of surface reconstruction in DTU from 3 views. \textbf{SparseNeuS and VolRecon use deep data-driven priors, whereas we do not}.
    }
    \label{fig:comparison_dtu}
\end{figure}

\begin{figure}[t!]
\begin{minipage}{0.50\linewidth}
\centering
\includegraphics[width=0.9\linewidth]{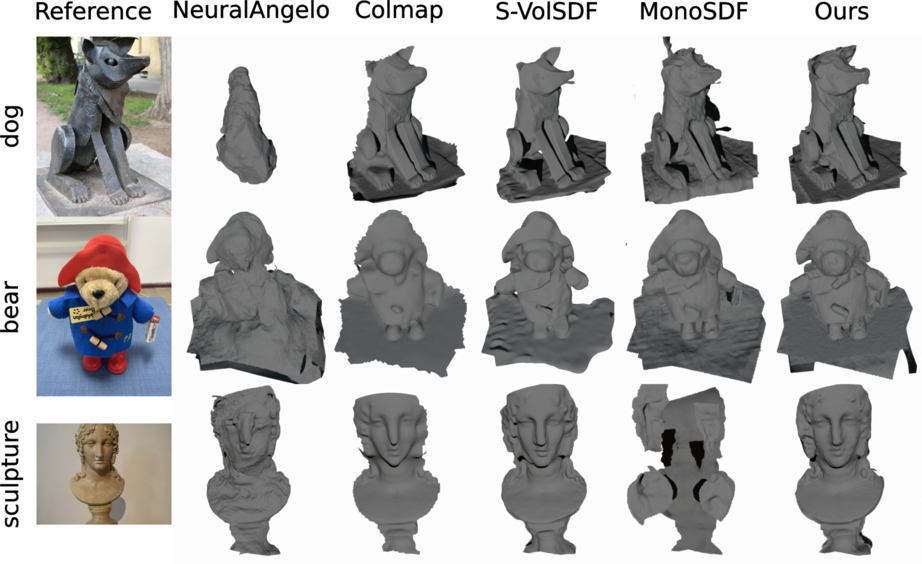}
\caption{Qualitative comparison of surface reconstruction in BMVS from 3 views. 
}
    \label{fig:comparison_bmvs}

    \end{minipage}%
    \hfill
\begin{minipage}{0.45\linewidth}
\centering
    \includegraphics[width=1.0\linewidth]{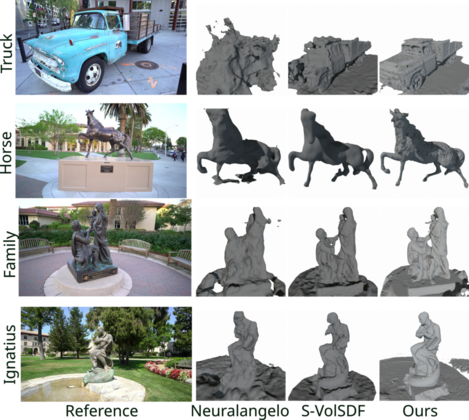}
    \caption{Qualitative comparison of surface reconstruction on T$\&$T from 24 uniformly sampled views.}
    \label{fig:comparison_tnt}
    \end{minipage}
\end{figure}

\begin{figure}[t!]
    \begin{minipage}{0.48\linewidth}
    \centering
    \includegraphics[width=1.0\linewidth]{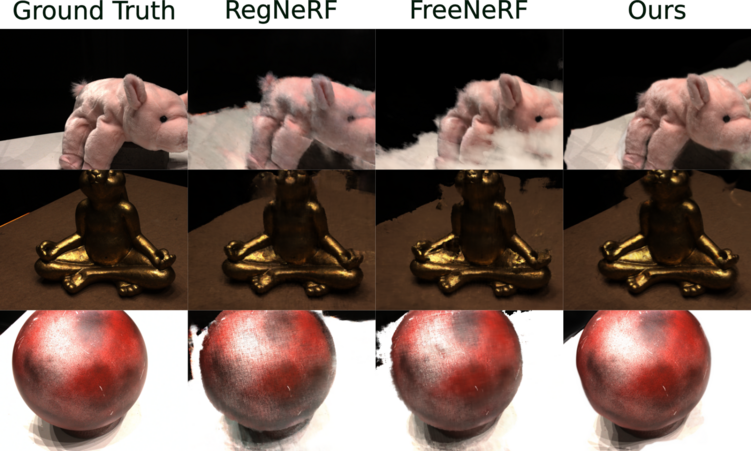}
    \caption{Qualitative comparison of novel view synthesis in DTU from 3 views. 
    }
    \label{fig:comparison_3v}
    \end{minipage}%
    \hfill
    \begin{minipage}{0.48\linewidth}
    \centering
    \includegraphics[width=1.0\linewidth]{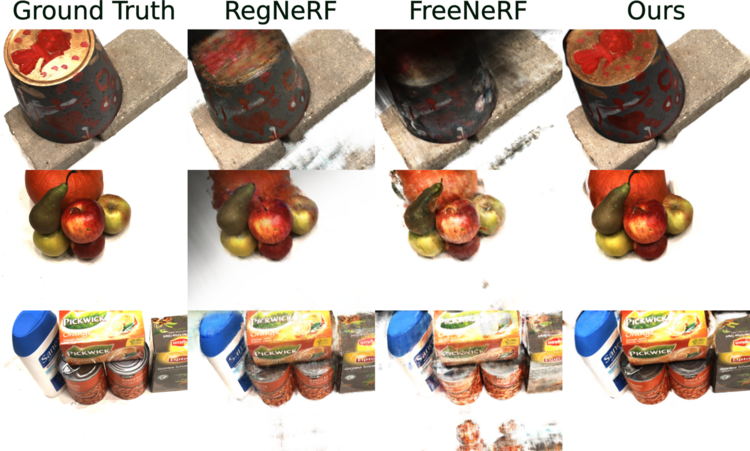}
    \caption{Qualitative comparison of novel view synthesis in DTU from 6 views. 
    }
    \label{fig:comparison_6v}
    \end{minipage}
\end{figure}

\noindent\subsection{Datasets and Setups}
We follow the experimental settings in RegNeRF~\cite{niemeyer2022regnerf} for novel view synthesis from sparse views (3, 6 and 9). We follow SparseNeuS~\cite{long2022sparseneus} for reconstruction from 3 views. More details can be found in the supplementary material. For all qualitative figures, more examples are provided in the supplementary material.

\noindent\textbf{Few-Shot Novel View Synthesis}
We evaluate on 15 scenes from the DTU dataset~\cite{jensen2014large}. We follow the protocol in sparse NeRF-based methods~\cite{niemeyer2022regnerf, yu2021pixelnerf, yang2023freenerf, Seo_2023_ICCV}. Following these methods, we report here the foreground metric, and provide the full image metric in the supplementary material as-well. We report PSNR, SSIM, VGG LPIPS scores, and the geometric average, following~\cite{niemeyer2022regnerf}.

We relay results reported by~\cite{yang2023freenerf, Seo_2023_CVPR, wynn2023diffusionerf}. We mainly compare against state-of-the-art generalizable methods PixelNeRF~\cite{yu2021pixelnerf}, Stereo Radiance Fields (SRF)~\cite{chibane2021stereo} and MVSNeRF~\cite{chen2021mvsnerf} as pretrained and fine-tuned (denoted with "ft") by RegNeRF~\cite{niemeyer2022regnerf}. We also compare against NeRF-based methods that use external priors ~\cite{niemeyer2022regnerf,jain2021putting,wynn2023diffusionerf}, as well as NeRF-based regularization methods~\cite{yang2023freenerf,Seo_2023_ICCV,Seo_2023_CVPR}. 

\begin{table}[t]
\begin{center}
\scalebox{0.7}{
\begin{tabular}{l|lllllllllllllll|l}
%\cline{17-17}
\textbf{\textbf{Scan}}                     & 24            & 37            & 40            & 55            & 63            & 65            & 69            & 83            & 97           & 105           & 106           & 110           & 114           & 118           & 122           & Mean $\downarrow$          \\ \hline %\cline{17-17} 
COLMAP~\cite{schonberger2016structure}     & \textbf{0.9}  & 2.89          & 1.63          & 1.08          & 2.18          & 1.94          & 1.61          & \textit{1.3}  & 2.34         & 1.28          & 1.1           & 1.42          & 0.76          & 1.17          & \textit{1.14} & 1.52          \\ %\cline{17-17} 
IDR~\cite{idr_yariv2020multiview}              & 4.01          & 6.4           & 3.52          & 1.91          & 3.96          & 2.36          & 4.85          & 1.62          & 6.37         & 5.97          & 1.23          & 4.73          & 0.91          & 1.72          & 1.26          & 3.39          \\ %\cline{17-17} 
VolSDF~\cite{yariv2021volume}              & 4.03          & 4.21          & 6.12          & \textit{0.91} & 8.24          & \textit{1.73} & 2.74          & 1.82          & 5.14         & 3.09          & 2.08          & 4.81          & 0.6           & 3.51          & 2.18          & 3.41          \\ %\cline{17-17} 
UNISURF~\cite{oechsle2021unisurf}          & 5.08          & 7.18          & 3.96          & 5.3           & 4.61          & 2.24          & 3.94          & 3.14          & 5.63         & 3.4           & 5.09          & 6.38          & 2.98          & 4.05          & 2.81          & 4.39          \\ %\cline{17-17} 
NeuS~\cite{wang2021neus}                   & 4.57          & 4.49          & 3.97          & 4.32          & 4.63          & 1.95          & 4.68          & 3.83          & 4.15         & 2.5           & 1.52          & 6.47          & 1.26          & 5.57          & 6.11          & 4.00          \\ %\cline{17-17} 
SparseNeuS ft~\cite{long2022sparseneus}    & 1.29          & \underline{2.27}    & \textit{1.57} & \underline{0.88}    & 1.61          & 1.86          & \textit{1.06} & \underline{1.27}    & \underline{1.42}   & 1.07          & \underline{0.99}    & \textit{0.87} & \underline{0.54}    & \textit{1.15} & 1.18          & \textit{1.27} \\ %\cline{17-17} 
MVSNeRF~\cite{chen2021mvsnerf}                     & 1.96          & 3.27          & 2.54          & 1.93          & 2.57          & 2.71          & 1.82          & 1.72          & 2.29         & 1.75          & 1.72          & 1.47          & 1.29          & 2.09          & 2.26          & 2.09          \\ %\cline{17-17} 
PixelNerf~\cite{yu2021pixelnerf}           & 5.13          & 8.07          & 5.85          & 4.4           & 7.11          & 4.64          & 5.68          & 6.76          & 9.05         & 6.11          & 3.95          & 5.92          & 6.26          & 6.89          & 6.93          & 6.28          \\ %\cline{17-17} 
SparseNeuS infer~\cite{long2022sparseneus} & 1.68          & 3.06          & 2.25          & 1.1           & 2.37          & 2.18          & 1.28          & 1.47          & 1.8          & 1.23          & 1.19          & 1.17          & 0.75          & 1.56          & 1.55          & 1.64          \\ %\cline{17-17} 
VolRecon~\cite{ren2022volrecon}            & 1.2           & 2.59          & \underline{1.56}    & 1.08          & \textit{1.43} & 1.92          & 1.11          & 1.48          & \underline{1.42}   & \textit{1.05} & 1.19          & 1.38          & 0.74          & 1.23          & 1.27          & 1.38          \\ %\cline{17-17} 
ReTR~\cite{liang2023retr}                  & \underline{1.05}    & \textit{2.31} & \textbf{1.44} & 0.98          & \textbf{1.18} & \textbf{1.52} & \underline{0.88}    & 1.35          & \textbf{1.3} & \underline{0.87}    & \textit{1.07} & \underline{0.77}    & \textit{0.59} & \underline{1.05}    & \underline{1.12}    & \underline{1.17}    \\ %\cline{17-17} 
Ours (SparseCraft)                       & \textit{1.17} & \textbf{1.74} & 1.8           & \textbf{0.7}  & \underline{1.19}    & \underline{1.53}    & \textbf{0.83} & \textbf{1.05} & \underline{1.42}   & \textbf{0.78} & \textbf{0.8}  & \textbf{0.56} & \textbf{0.44} & \textbf{0.77} & \textbf{0.84} & \textbf{1.04} \\ %\cline{17-17} 
\end{tabular}}
\caption{Quantitative results of sparse view surface reconstruction on 15 testing scenes of DTU dataset~\cite{jensen2014large}. We report Chamfer distance (lower is better). Best scores are in \textbf{bold}, second best are \underline{underlined} and third best are in \textit{italic}.}
\label{tab:reconstruction_comparison}
\end{center}
\end{table}

\begin{table}[t]
\begin{center}
\scalebox{0.7}{
\begin{tabular}{l|lll|lll|lll|lll}

&\multicolumn{3}{c}{Object PSNR $\uparrow$}                  & \multicolumn{3}{|c}{Object SSIM $\uparrow$}                  & \multicolumn{3}{|c}{Object LPIPS $\downarrow$}                 & \multicolumn{3}{|c}{Object Average $\downarrow$}               \\
                                         & 3 views        & 6 views        & 9 views        & 3 views        & 6 views        & 9 views        & 3 views        & 6 views        & 9 views        & 3 views        & 6 views        & 9 views        \\
\hline                                         
SRF~\cite{chibane2021stereo}             & 15.32          & 17.54          & 18.35          & 0.671          & 0.73           & 0.752          & 0.304          & 0.25           & 0.232          & 0.171          & 0.132          & 0.12           \\
PixelNeRF~\cite{yu2021pixelnerf}         & 16.82          & 19.11          & 20.4           & 0.695          & 0.745          & 0.768          & 0.27           & 0.232          & 0.22           & 0.147          & 0.115          & 0.1            \\
MVSNerf~\cite{chen2021mvsnerf}                   & 18.63          & 20.7           & 22.4           & \textit{0.769} & 0.823          & 0.853          & 0.197          & 0.156          & 0.135          & 0.113          & 0.088          & 0.068          \\
SRF ft~\cite{chibane2021stereo}          & 15.68          & 18.87          & 20.75          & 0.698          & 0.757          & 0.785          & 0.281          & 0.225          & 0.205          & 0.162          & 0.114          & 0.093          \\
PixelNeRF ft~\cite{yu2021pixelnerf}      & 18.95          & 20.56          & 21.83          & 0.71           & 0.753          & 0.781          & 0.269          & 0.223          & 0.203          & 0.125          & 0.104          & 0.09           \\
MVSNeRF ft~\cite{chen2021mvsnerf}                & 18.54          & 20.49          & 22.22          & \textit{0.769} & 0.822          & 0.853          & 0.197          & 0.155          & 0.135          & 0.113          & 0.089          & 0.069          \\
DietNeRF~\cite{jain2021putting}          & 11.85          & 20.63          & 23.83          & 0.633          & 0.778          & 0.823          & 0.314          & 0.201          & 0.173          & 0.243          & 0.101          & 0.068          \\
RegNerf~\cite{niemeyer2022regnerf}       & 18.89          & 22.2           & 24.93          & 0.745          & \textit{0.841} & \textit{0.884} & 0.19           & 0.117          & 0.089          & 0.112          & 0.071          & 0.047          \\
FreeNerf~\cite{yang2023freenerf}         & \underline{19.92}    & \underline{23.25}    & \underline{25.38}    & \underline{0.787}    & \underline{0.844}    & \underline{0.888}    & \textit{0.182} & 0.137          & 0.096          & \underline{0.098}    & 0.068          & 0.046          \\
MixNerf~\cite{Seo_2023_CVPR}             & 18.95          & 22.3           & 25.03          & 0.744          & 0.835          & 0.879          & 0.203          & \textit{0.102} & \textit{0.065} & 0.113          & \textit{0.066} & \textit{0.042} \\
FlipNerf~\cite{Seo_2023_ICCV}            & \textit{19.55} & \textit{22.45} & 25.12          & 0.767          & 0.839          & 0.882          & \underline{0.18}     & \underline{0.098}    & \underline{0.062}    & \textit{0.101} & \underline{0.064}    & \underline{0.041}    \\
DiffusioNerf~\cite{wynn2023diffusionerf} & 16.2           & 20.34          & \textit{25.18} & 0.698          & 0.818          & 0.883          & 0.207          & 0.139          & 0.095          & 0.146          & 0.081          & 0.047          \\
Ours (SparseCraft)                    & \textbf{20.55} & \textbf{23.72} & \textbf{26.03} & \textbf{0.832} & \textbf{0.888} & \textbf{0.917} & \textbf{0.116} & \textbf{0.074} & \textbf{0.058} & \textbf{0.084} & \textbf{0.052} & \textbf{0.037}
\end{tabular}}
\caption{Quantitative comparison on DTU. We present the PSNR, SSIM, VGG LPIPS and Average scores of foreground objects. Best scores are in \textbf{bold}, second best are \underline{underlined} and third best are in \textit{italic}.}
\label{tab:novelview_comparison} 
\end{center}
%\vspace{-45pt}
\end{table}

\noindent\textbf{Few-Shot Reconstruction}
We evaluate on datasets DTU~\cite{jensen2014large}, BlendedMVS~\cite{yao2020blendedmvs} and Tanks \& Temples~\cite{knapitsch2017tanks}. We use the same 15 testing scenes as SparseNeuS~\cite{long2022sparseneus} (Please note that the DTU splits and views used for novel view synthesis and reconstruction are not the same).
Each scene is evaluated on two sets of 3 different views. We use Chamfer distance as metric and report the average of the two sets for each scene, similarly to~\cite{long2022sparseneus, ren2022volrecon}. 
We use the same evaluation script as this benchmark, \ie cleaning the generated meshes with masks of training views and sampling points from the generated meshes. We further test our method on few challenging scenes from BlendedMVS~\cite{yao2020blendedmvs}. We also evaluate our method on large-scale scenes from Tanks \& Temples dataset~\cite{knapitsch2017tanks} using only 24 views from the total of more than 150.

For DTU, we report the evaluation as in~\cite{ren2022volrecon, liang2023retr}. 
We compare mainly against the previously introduced conditional models~\cite{yu2021pixelnerf,chen2021mvsnerf}, generalizable reconstruction methods~\cite{ren2022volrecon, liang2023retr,long2022sparseneus} as well as per-scene optimization based neural surface reconstruction methods~\cite{idr_yariv2020multiview,yariv2021volume,oechsle2021unisurf,wang2021neus} and the fine-tuned SparseNeuS~\cite{long2022sparseneus} (denoted SparseNeuS ft). We note that the generalizable methods VolRecon~\cite{ren2022volrecon} and ReTR~\cite{liang2023retr} do not allow per-scene fine-tuning. We also compare against COLMAP~\cite{schoenberger2016mvs}.

As there is no standard benchmark for BlendedMVS~\cite{yao2020blendedmvs}, we use it for qualitative evaluation. We compare against MonoSDF~\cite{yu2022monosdf}, that uses monocular depth and normal priors, COLMAP~\cite{schoenberger2016mvs},
NeuralAngelo~\cite{li2023neuralangelo} the state-of-the-art hash-based reconstruction method in the dense setting,
and S-VolSDF~\cite{wu2023s}, a method that improves the performance of deep MVS through the volumetric rendering of VolSDF~\cite{yariv2021volume}. 

We also compare our method qualitatively on the large-scale dataset of Tanks \& Temples~\cite{knapitsch2017tanks} against 
data prior based and test-time optimization method S-VolSDF~\cite{wu2023s}, and NeuralAngelo~\cite{li2023neuralangelo}.
\subsection{Surface Reconstruction}

As shown in Table~\ref{tab:reconstruction_comparison}, our method SparseCraft outperforms the SOTA on average and on most scenes, even against the generalizable models that were pretrained on other scenes of the same datasets. In particular, we show substantial improvement for challenging scenes with shiny objects such as scans 110 and 37 as shown in the qualitative comparison. This showcases also that our method improves largely on the leveraged result from MVS (COLMAP~\cite{schoenberger2016mvs}), as the latter is known for struggling with shiny/reflective surfaces.

As for qualitative visualizations on both DTU~\cite{jensen2014large} and BlendedMVS~\cite{yao2020blendedmvs} presented in Figures~\ref{fig:comparison_dtu} and~\ref{fig:comparison_bmvs}, our method generates overall more detailed and complete surfaces compared to previous methods. For instance, for scan 118 from DTU~\cite{jensen2014large}, NeuS~\cite{wang2021neus} generates an inaccurate surface with many wholes. The fine-tuned SparseNeuS~\cite{long2022sparseneus} generates a relatively complete surface, but overly smooth and lacking important details. VolRecon~\cite{ren2022volrecon} displays more details compared to SparseNeuS-ft, but the surface normals appear to be noisy and inaccurate.
Our method can achieve such performance while being faster than the per-scene optimization methods (Table~\ref{tab:time}). We note that obtaining the MVS point cloud (COLMAP) takes only $41$ seconds in 3 views, $180$ seconds for 9 views in DTU.
Our surfaces show more fidelity and a better trade-off between details and smoothness.
On the challenging large scale dataset T\&T (Figure \ref{fig:comparison_tnt}), we found that the generalizable VolRecon~\cite{ren2022volrecon} fails to generate reasonable outputs. 
Notice that our reconstructions display more fidelity and details and fewer failures in this large scale setting, even-though only a limited number of views is used.

\subsection{Novel View Synthesis}

As shown in table~\ref{tab:novelview_comparison}, for all input setting, we outperform the current SOTA by a large margin in all metrics, especially in the most extreme case of 3 input views. In fact, our method shows superior results on VGG LPIPS which is reflected in the qualitative comparison in the 3 and 6 input-view settings~\ref{fig:comparison_3v} and~\ref{fig:comparison_6v} respectively, where our renderings appear to be more photo-realistic compared to RegNeRF~\cite{niemeyer2022regnerf} and FreeNeRF~\cite{yang2023freenerf}. For example, the red ball and shiny golden rabbit show how our method handles well challenging light reflections, and the example of colorful fruits shows how our method can handle high dynamic range, all thanks to the various considerations in the design of our method, as well as the proposed regularization for the sparse setting. 

\section{Ablation}
\label{sec:ablation}

\begin{figure}[t!]
\begin{minipage}{0.47\linewidth}
\centering
\scalebox{0.85}{
\begin{tabular}{l|l}
Applied losses      & Chamfer Distance $\downarrow$ \\ \hline
SDF loss  & 1.34             \\
$\mathcal{L_{\text{tay}}}(\mathbf{p})$  & 1.17             \\
$\mathcal{L_{\text{tay}}}(\mathbf{p})$ + Normal loss & 1.10             \\
$\mathcal{L_{\text{tay}}}(\mathbf{p})$ + $\mathcal{L_{\text{tay}}}(\mathbf{q})$  & \textbf{1.08}   
\end{tabular}
}
\label{tab:fs}
\captionof{table}{Numerical Ablation of our Taylor based geometric regularization losses.}
\label{tab:taylor_chamfer}
\begin{tabular}{lllll}
\hline
All losses &       & $\times$     &       & $\times$             \\
Prog. Enc. &       &       & $\times$     & $\times$              \\ \hline
Chamfer $\downarrow$   & 4.11 & 1.16 & 2.56 & \textbf{1.01} \\
PSNR $\uparrow$  & 15.65 & 16.14 & 18.06 & \textbf{20.55}
\end{tabular}

\captionof{table}{Numerical ablation of progressive hash encoding, and the MVS based regularization losses.}
\label{tab:prog_psnr} 
       
    \end{minipage}
    \hfill
\begin{minipage}{0.50\linewidth}
\centering
    \includegraphics[width=1.0\linewidth]{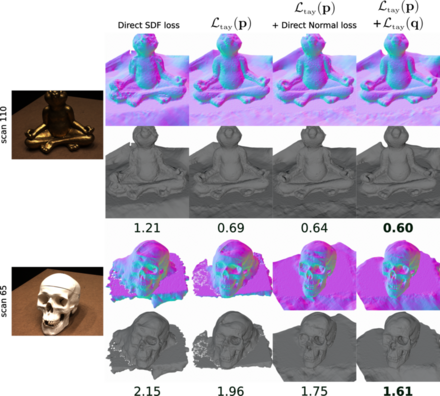}
    \caption{Ablation of our Taylor based geometric regularization losses. We report the \textbf{Chamfer score} of reconstructions.}
    \label{fig:taylor_ablation}
    \end{minipage}
\end{figure}

For all reconstruction experiments henceforth, we show average performance over one of the two sets of 3 views of all 15 DTU scenes. For novel view synthesis experiments, we report average metrics on all 15 scenes of DTU. We note that besides the ablations presented below, additional ablations are provided in the supplementary, including the influence of MVS point cloud's density, the performance when varying the number of input views, several studies on design choices related to the sampling of Taylor query points and the effect of progressive hash encoding scheduling.

\noindent\textbf{Taylor losses \vs Direct MVS supervision} We compare our Taylor expansion based geometric losses to their baselines. The input Taylor loss $\mathcal{L}_{\text{tay}}(\mathbf{p})$ baseline is direct SDF zero supervision. The query Taylor loss  $\mathcal{L}_{\text{tay}}(\mathbf{q})$ baseline is a direct SDF gradient supervision with the normal $\mathbf{n}_{\text{\tiny MVS}}(\mathbf{p})$. As can be seen in Figure~\ref{fig:taylor_ablation} and Table~\ref{tab:taylor_chamfer}, our proposed losses outperform the other combinations and displays improved details and less noise in the reconstructions. This can also be witnessed by the chamfer scores reported in Figure~\ref{fig:taylor_ablation}. While direct MVS supervision, especially the normal loss, improves our baseline, incorporating this supervision through our Taylor losses is more beneficial. We also find that the proposed Input Taylor loss is largely superior to applying only the SDF zero supervision, which validate our hypothesis about the benefit of enforcing linearity close to surface points.

\noindent\textbf{Ablation of progressive hash encoding} Table~\ref{tab:prog_psnr} shows improvement brought by the progressive hash encoding, and the MVS based regularization losses. We find that while our regularization can improve surface quality, using progressive encoding act as a regularization and helps to avoid artifacts in the reconstruction, so that the geometry model does not prematurely overfit to fine details. In addition, as our proposed losses are geometric in nature, they sacrifice rendering quality at the expense of good reconstruction. Combining them with the progressive encoding leads to superior rendering quality than using only the progressive encoding.

\begin{figure}[ht!]
\begin{minipage}{0.60\linewidth}
\centering
    \includegraphics[width=0.9\linewidth]{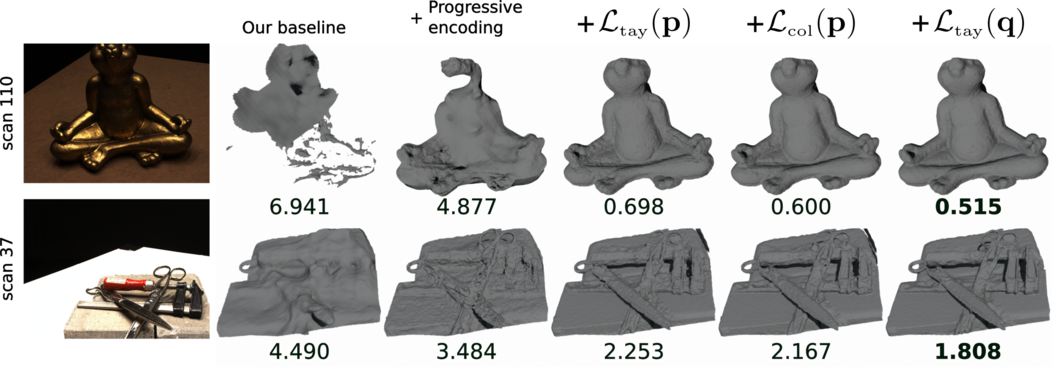}
\caption{Ablation of our method's components. \\\hspace{\textwidth}
We report the \textbf{Chamfer score} of reconstructions.}
    \label{fig:ablation_chamfer}
    \end{minipage}%
    \hfill
\begin{minipage}{0.40\linewidth}
\centering
\scalebox{0.7}{
\begin{tabular}{llllllll}
\hline
$\mathcal{L_{\text{tay}}}(\mathbf{p})$ & $\times$            & $\times$    & $\times$   & $\times$    &      &      &      \\
$\mathcal{L_{\text{tay}}}(\mathbf{q})$ & $\times$             & $\times$    &      &      & $\times$    &      &      \\
$\mathcal{L_{\text{col}}}(\mathbf{p})$    & $\times$             &      & $\times$    &      &      & $\times$    &      \\ \hline
Chamfer  $\downarrow$ & \textbf{1.01} & 1.08 & 1.09 & 1.17 & 1.14 & 2.41 & 4.11 \\
PSNR $\uparrow$      & \textbf{20.55} & 19.65 & 19.44 & 19.27 & 20.12 & 18.05 & 15.65 \\ 
LPIPS $\downarrow$      & \textbf{0.116} & 0.132 & 0.146 & 0.152 & 0.127 & 0.192 & 0.269
\end{tabular}}
\captionof{table}{Numerical ablation of our MVS based regularization losses.}
\label{tab:ablation_chamfer} 
    \end{minipage}
\end{figure}

\noindent\textbf{Ablation of regularization losses} Table \ref{tab:ablation_chamfer} and Figure \ref{tab:ablation_chamfer} illustrate the contribution of each of our regularization losses to our final performance for both reconstruction and rendering quality. Our baseline model in this case is our method without progressive encoding and MVS regularization. We find that both Taylor-based losses are crucial for learning good surfaces. Further, regularizing the diffuse component of the color network,  with MVS color labels alleviates the issue of bias found in the rendering process of NeuS as studied in~\cite{fu2022geoneus}, and hence improves the performance as well while enhancing rendering results even further.   

\begin{figure}[t!]
\begin{minipage}{0.48\textwidth}
\centering
 \includegraphics[width=0.8\linewidth]{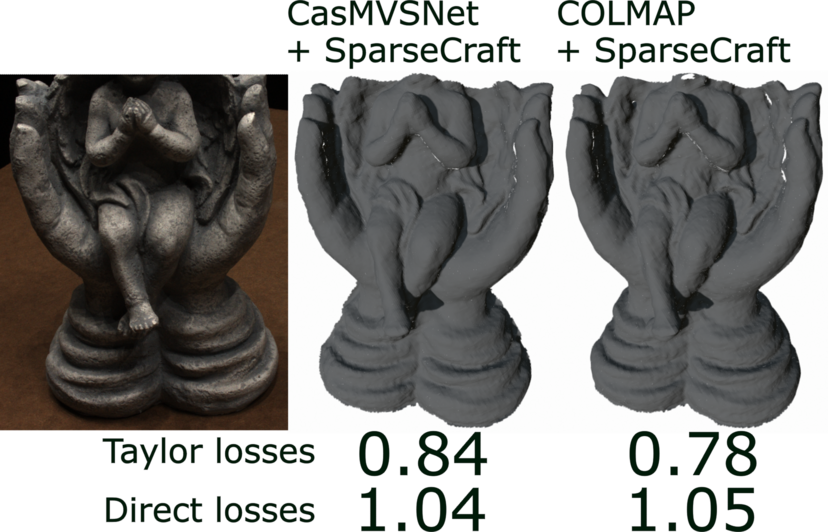}
        \caption{Running our method with learnable MVS. We report the \textbf{Chamfer score} of reconstruction.}
        \label{fig:mvsnet}

\end{minipage}
\hfill
\begin{minipage}{0.48\textwidth}
\centering
\scalebox{0.7}{
\begin{tabular}{l|l}
Method                                  & Training Time $\downarrow$ \\ \hline
Ours (SparseCraft)                          & 9 minutes                                \\
Ours (SparseCraft) w/o reg.                  & 7 minutes                                \\
SparseNeuS ft~\cite{long2022sparseneus} & 20 minutes                               \\
NeuralAngelo~\cite{li2023neuralangelo}  & 15 minutes                               \\
S-VolSDF~\cite{wu2023s}                 & 18 minutes                               \\
MonoSDF~\cite{yu2022monosdf}            & 1.5 hours+                              
\end{tabular}}
\captionof{table}{Training time on an NVIDIA RTX A6000 of per-scene optimization methods for surface reconstruction from 3 views.} 
\label{tab:time}
\end{minipage}
\end{figure}

\noindent\textbf{Using learnable MVS} Figure~\ref{fig:mvsnet} shows the compatibility of our method with other Point Clouds sources, in this case CasMVSNet\cite{gu2020cascade}. Notice that our novel Taylor losses improve over standard direct losses both when using COLMAP and CasMVSNet.

\section{Limitations}
\label{sec:limit}
Since our method uses MVS cues, it suffers from the same limitations of the used MVS method (COLMAP in our case); Thus, it requires enough overlap between input images, and it may not be suitable for reconstruction of highly non-Lambertian surfaces, for which COLMAP is known to fail. In addition, The point cloud obtained from the MVS method have to be dense enough for more accurate normals estimation used by our method. We showed in our experiments how these limitations could be alleviated to some extent by using more advanced MVS techniques such as learnable MVS.

\section{Conclusion}
\label{sec:conc}
We presented a new method called SparseCraft for time efficient learning of SDF and radiance fields from sparse imagery. We bridged photogrammetry and deep learning based INRs through novel regularization losses to obtain the SOTA in novel view synthesis and reconstruction simultaneously. Through input data requirement reduction, we hope this work will contribute towards more accessible 3D capture. 

\bibliographystyle{splncs04}
\bibliography{main}
\end{document}